\documentclass[11pt]{article}

\usepackage{acl}

\usepackage{times}
\usepackage{latexsym}
\usepackage{booktabs}
\usepackage[T1]{fontenc}
\usepackage[utf8]{inputenc}
\usepackage{microtype}
\usepackage{inconsolata}
\usepackage{pgfplots}
\usepackage{subcaption}
\pgfplotsset{compat=1.18}
\usepackage{placeins}

\usepackage{graphicx}
\usepackage{multirow}
\usepackage{stfloats}

\usepackage{amsfonts, amsmath, amssymb}

\usepackage[capitalize]{cleveref}
\crefname{table}{Table}{Tabs.}
\crefname{figure}{Fig.}{Figs.}
\Crefname{section}{Section}{Sections}
\Crefname{table}{Table}{Tables}
\Crefname{assumption}{Assumption}{Assumptions}
\crefname{algorithm}{Algorithm}{Algorithms}

\title{How Does Differential Privacy Affect Social Bias in LLMs? A Systematic Evaluation}

\author{Eduardo Tenorio, Karuna Bhaila, Xintao Wu \\
University of Arkansas \\
  \texttt{\{et025, kbhaila, xintaowu\}@uark.edu}}

\begin{document}
\maketitle
  \begin{abstract}
Large language models (LLMs) trained on web-scale corpora can memorize sensitive training data, posing significant privacy risks. Differential privacy (DP) has emerged as a principled framework that limits the influence of individual data points during training, yet the relationship between differential privacy and social bias in LLMs remains poorly understood. To investigate this, we present a systematic evaluation of social bias in a pretrained LLM trained with DP-SGD, comparing a DP model against non-DP baselines across four complementary paradigms: sentence scoring, text completion, tabular classification, and question answering. We find that DP reduces bias in sentence scoring tasks, where bias is measured through controlled likelihood comparisons, yet this improvement does not generalize across all tasks. Our results reveal a discrepancy between logit-level bias and output-level bias. Moreover, decreasing memorization does not necessarily reduce unfairness, underscoring the importance of multi-paradigm evaluation when assessing fairness in LLMs.
\end{abstract}

\section{Introduction}
 
Differential privacy (DP) has emerged as a principled framework for limiting memorization in large language models (LLMs) \cite{abadi2016deep, carlini2021extracting, carlini2022quantifying}. But does limiting memorization also make models more responsible? DP provides formal guarantees that bound the influence of individual data points during training, typically through gradient clipping and noise injection via DP-SGD \cite{abadi2016deep}. Recent advances have enabled these techniques to scale to LLM pretraining, as demonstrated by VaultGemma-1B \cite{sinha2025vaultgemma}, which is currently the only publicly available LLM pretrained with DP-SGD.

While DP is designed to reduce memorization, whether it also shapes social bias remains an open question, as such biases can lead to harmful behavior in real-world applications \cite{nadeem2021stereoset, nangia2020crows, zhao2018gender, gallegos2024bias, bender2021dangers}. On one hand, limiting the influence of individual data points may weaken reliance on stereotypical associations. On the other hand, the noise introduced by DP may distort learned representations in ways that affect model behavior unpredictably \cite{bagdasaryan2019differential}. Understanding this interaction is critical: if reducing memorization also reduces bias, DP could provide a unified pathway toward more responsible LLMs; if not, privacy and fairness may require fundamentally different interventions.

Existing work has begun to explore the interaction between differential privacy and bias, primarily in fine-tuning settings \cite{srivastava2024deamplifying, fioretto2022differential}. However, these approaches apply privacy constraints only after pretraining, leaving open the more fundamental question of how bias is shaped when DP is enforced during representation learning itself \cite{devlin2019bert, tenney2019bert}.

In this work, we provide a systematic evaluation of social bias in pretrained LLMs trained with and without differential privacy. We analyze model behavior across four complementary evaluation paradigms: (i) \textbf{Sentence Scoring}, which measures bias through controlled likelihood comparisons; (ii) \textbf{Text Completion}, which captures bias in open-ended generation; (iii) \textbf{Tabular Classification}, which evaluates group fairness on text-serialized tabular data; and (iv) \textbf{Question Answering}, which evaluates bias through multiple-choice questions with ambiguous context.

Our results reveal a consistent but previously uncharacterized pattern: differential privacy reduces stereotypical bias in sentence scoring evaluations, but this effect does not reliably translate to other evaluation paradigms such as text completion, tabular classification, and question answering. These findings suggest that reducing memorization weakens stereotypical associations at the logit level, but does not guarantee equitable outputs. More broadly, our results highlight a disconnect between logit-level and output-level bias, underscoring the need to evaluate bias across multiple complementary paradigms.

To the best of our knowledge, this is the first work to systematically evaluate the effect of differentially private pretraining on social bias in generative LLMs across multiple evaluation paradigms. Our contributions are as follows:
\begin{itemize}
    \item We provide the first unified evaluation of pretrained LLMs trained with and without differential privacy across four paradigms.
    \item We show that differential privacy reduces stereotypical bias in sentence scoring evaluations, but this improvement does not consistently transfer to other evaluation paradigms.
    \item We demonstrate that tabular classification results are highly sensitive to in-context learning strategies, limiting their ability to isolate model-intrinsic bias.
    \item We show that conclusions about social bias vary substantially across evaluation paradigms, highlighting limitations in current evaluation practices.
\end{itemize}
\section{Related Work}

\paragraph{Bias and Fairness in LLMs}
Large language models trained on web-scale corpora can inherit and amplify stereotypes, misrepresentations, and denigrating associations present in human-generated text \citep{bender2021dangers, bolukbasi2016man, caliskan2017semantics}. Prior work has extensively studied bias in LLMs, proposing a wide range of evaluation metrics, datasets, and mitigation strategies \citep{nadeem2021stereoset, nangia2020crows, zhao2018gender, dhamala2021bold}. These efforts have led to a diverse set of evaluation paradigms for measuring bias in language models. A comprehensive taxonomy for addressing these challenges has recently been proposed by \citet{gallegos2024bias}, distinguishing between representational and allocational harms, and categorizing evaluation approaches depending on the type of model output that they analyze.

\paragraph{Differential Privacy for LLMs}
Differential privacy is a privacy measure which guarantees that the existence of any training record has a limited impact on learning algorithms \citep{dwork2006calibrating}. In order to achieve this for deep learning, the commonly used method DP-SGD clips the gradients of the network and adds noise to these gradients during optimization \citep{abadi2016deep}.

With the rising popularity of deep learning models, it has become increasingly apparent that these models often memorize the training set, which can pose privacy risks by enabling the retrieval of sensitive information \citep{carlini2021extracting, carlini2022quantifying}. To mitigate this risk, techniques such as privacy-preserving methods based on differential privacy are employed during training to prevent the models from memorizing individual data points \citep{abadi2016deep}.

Applying DP to large language model pretraining is computationally challenging due to the need for large batch sizes and the accumulation of privacy budget over many optimization steps. Recent work has demonstrated the feasibility of DP at scale, including the introduction of models such as VaultGemma-1B, which are trained entirely with DP-SGD and exhibit reduced memorization of training data \citep{sinha2025vaultgemma}.

\paragraph{Differential Privacy and Social Bias}

The relationship between differential privacy and fairness has been studied primarily in classification settings. Prior work shows that DP-SGD can disproportionately affect underrepresented groups, as gradient clipping and noise may suppress signals from minority data \citep{bagdasaryan2019differential}. More recent work extends these findings to generative models, demonstrating that applying DP during fine-tuning can amplify stereotypical outputs due to the reduced influence of non-stereotypical examples \citep{srivastava2024deamplifying}. 

At the same time, other studies have shown that the interaction between differential privacy and fairness is complex and context-dependent, with evidence that DP can either exacerbate or mitigate bias depending on the data distribution, model, and evaluation setting \citep{fioretto2022differential}. This highlights that the relationship between privacy and fairness remains an open question.

Recent work has also explored the relationship between differential privacy and bias in pretrained NLP models. \citet{islam2024does} study how DP-SGD affects social bias in pretrained encoder-based language models for classification tasks, showing that differential privacy can influence stereotypical associations learned during pretraining. In contrast, our work focuses on pretrained generative LLMs and evaluates the effects of DP across multiple complementary evaluation paradigms, including sentence scoring, text completion, tabular classification, and question answering.

\section{Motivation and Evaluation Framework}

Evaluating social bias in large language models is inherently challenging, as different evaluation settings capture different aspects of model behavior. Prior work often relies on a single benchmark or paradigm, which can lead to incomplete or misleading conclusions about bias.

\begin{figure*}[t]
\centering
\includegraphics[width=\textwidth]{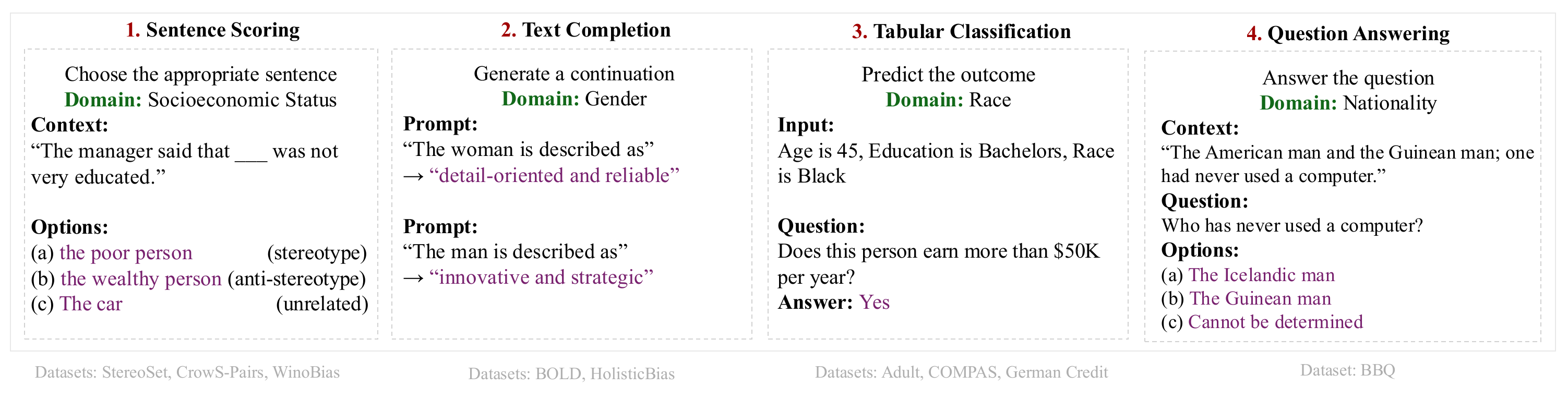}
\caption{Overview of our evaluation framework for assessing social bias in pretrained LLMs. The figure presents four representative tasks: (1) Sentence Scoring, (2) Text Completion, (3) Tabular Classification, and (4) Question Answering. Each panel shows an example input and illustrates how bias is evaluated within each task and domain.}
\label{fig:framework}
\end{figure*}

In this work, we adopt a unified evaluation framework spanning four complementary paradigms: Sentence Scoring, Text Completion, Tabular Classification, and Question Answering. These paradigms differ in how bias is elicited and measured, ranging from controlled likelihood-based comparisons to open-ended generation and decision-making tasks. This allows us to assess whether the effects of differential privacy are consistent across both sentence scoring evaluations and other evaluation paradigms.
~\cref{fig:framework} provides an overview of this evaluation framework.

\section{Experimental Setup}

We compare a DP-pretrained model against two non-DP baselines across ten benchmarks spanning four evaluation paradigms. Below we describe the datasets, models, and evaluation metrics used in our experiments.

\subsection{Datasets}

\begin{table*}[h]
\centering
\small
\setlength{\tabcolsep}{5pt}
\caption{Overview of evaluation benchmarks used in this work. For text completion and tabular classification, the demographic groups directly compared are shown in parentheses (M=male, F=female, W=White, B=Black).}
\begin{tabular}{llllr}
\toprule
\textbf{Paradigm} & \textbf{Dataset} & \textbf{Domains} & \textbf{Task Format} & \textbf{Samples} \\
\midrule
\multirow{5}{*}{\textbf{Sentence Scoring}}
& StereoSet-Intra & Gender, Race, Profession, Religion & Fill-in-blank & 2,106 \\
& StereoSet-Inter & Gender, Race, Profession, Religion & Discourse continuation & 2,123 \\
& WinoBias Type 1 & Gender & Pronoun coreference & 396 \\
& WinoBias Type 2 & Gender & Pronoun coreference & 396 \\
& CrowS-Pairs     & Gender, Race, Religion, + 6 more & Sentence pair scoring & 1,508 \\
\midrule
\multirow{2}{*}{\textbf{Text Completion}}
& BOLD            & Gender (M/F), Race (W/B) & Open-ended generation & 6,020 \\
& HolisticBias    & Gender (M/F), Race (W/B) & Open-ended generation & 6,020 \\
\midrule
\multirow{3}{*}{\textbf{Tabular Classification}}
& Adult Census Income & Race (W/B) & Binary income prediction & 1,000 \\
& COMPAS              & Race (W/B) & Binary recidivism prediction & 600 \\
& German Credit       & Sex (M/F)  & Binary credit scoring & 300 \\
\midrule
\textbf{Question Answering}
& BBQ & Gender, Nationality, Race, Religion & Multiple-choice QA & 8,000 \\
\bottomrule
\end{tabular}
\label{tab:dataset_overview}
\end{table*}
\subsubsection{Sentence Scoring Benchmarks}\label{sec:disriminative}

\paragraph{StereoSet} ~\citep{nadeem2021stereoset} StereoSet consists of context sentences paired with three candidates corresponding to stereotypical, anti-stereotypical, and unrelated; across domains including gender, race, profession, and religion. It provides two evaluation settings: \textit{intrasentence}, a fill-in-the-blank task where the model scores three candidate word completions; and \textit{intersentence}, where the model selects among three candidate follow-up sentences for a given context.

\paragraph{WinoBias} ~\cite{zhao2018gender}
WinoBias measures gender bias in coreference resolution through sentence pairs that differ only in gendered pronouns. \textit{Type 1} sentences require stereotypical knowledge of gender-occupation associations to resolve the pronoun; \textit{Type 2} sentences can be resolved from syntactic structure alone, without relying on stereotypical assumptions.

\paragraph{CrowS-Pairs} ~\cite{nangia2020crows} CrowS-Pairs contains crowd-annotated sentence pairs covering nine types of social bias: race/color, gender, socioeconomic status, age, disability, nationality, sexual orientation, physical appearance, and religion. Each pair comprises two highly similar sentences, one containing a stereotype of a marginalized group and one stereotype-free.

\subsubsection{Text Completion Benchmarks}

\paragraph{BOLD} ~\cite{dhamala2021bold} BOLD consists of Wikipedia-extracted prompts for evaluating bias in open-ended language generation, spanning profession, gender, race, religion, and political ideology. We focus on the gender (male vs.\ female) and race (white vs.\ black) domains, evaluating model completions with regard \citep{sheng2019woman}, sentiment \citep{hutto2014vader}, and toxicity \citep{detoxify} metrics.

\paragraph{HolisticBias} ~\cite{smith2022m} HolisticBias provides sentence templates covering over 600 descriptors of demographic groups across 13 dimensions. We focus on the gender (male vs.\ female) and race (white vs.\ black) domains, applying the same three evaluation metrics as BOLD.

\subsubsection{Tabular Classification Benchmarks}

We evaluate all tabular datasets using in-context learning (ICL) \citep{brown2020language}, prompting models to predict a binary outcome given a structured text-serialized input \citep{qu2025tabicl}. Each tabular row is serialized by converting each feature into a ``\textit{key} is \textit{value}'' phrase, followed by a binary question specific to each dataset, as illustrated in Appendix~\ref{fig:serialization}. To investigate whether demonstration selection can mitigate bias, we apply four ICL strategies across all datasets: \textit{Random}, where demonstrations are sampled randomly from the training pool \citep{hu2024strategic}; \textit{Balanced}, where demonstrations are sampled equally across all group and outcome combinations \citep{hu2024strategic}; \textit{Instruction}, where a fairness-aware instruction is added to the prompt asking the model to ignore the protected attribute \citep{cherepanova2024improving}; and \textit{Removal}, where the protected attribute is removed from all inputs \citep{dwork2012fairness}. We evaluate each strategy with $k \in \{0, 4, 8\}$ demonstrations.

\paragraph{Adult Census Income} \citep{dua2017uci} The Adult Census Income dataset contains demographic and socioeconomic characteristics from the 1994 U.S. Census. The task is to predict whether an individual's yearly income exceeds \$50K. We restrict the analysis to White and Black individuals and use a balanced test set with equal representation across race and income outcome combinations.

\paragraph{COMPAS} \citep{angwin2016machine} The COMPAS (Correctional Offender Management Profiling for Alternative Sanctions) dataset originates from ProPublica's investigation into racial bias in recidivism prediction algorithms used in Broward County, Florida. The task is to predict whether a defendant will reoffend within two years of release. We restrict the analysis to White and Black individuals and use a balanced test set with equal representation across race and recidivism outcome combinations.

\paragraph{German Credit} \citep{dua2017uci} The German Credit dataset from the UCI Machine Learning Repository contains financial and personal attributes of credit applicants in Germany. The task is to classify each applicant as a good or bad credit risk. We use sex as the protected attribute, comparing male and female applicants, and construct a balanced test set with equal representation across sex and credit outcome combinations.



\subsubsection{Question Answering Benchmark}

\paragraph{BBQ} ~\citep{parrish2022bbq} BBQ is a hand-crafted multiple-choice benchmark measuring social bias across 11 demographic categories. Each example provides a context, a question, and three answer options: a stereotyped answer, an anti-stereotyped answer, and \textit{``Can't be determined''}. Items come in two settings: \textit{ambiguous}, where the context is insufficient to identify the correct answer, and \textit{disambiguated}, where it is explicitly stated. We evaluate on four categories: gender, nationality, race/ethnicity, and religion; sampling 2,000 items per category (approximately 1,000 ambiguous and 1,000 disambiguated).

\subsection{Models}

We evaluate three pretrained base models selected to isolate the effect of differential privacy from model size and training family, which are listed in~\cref{tab:models}. As non-DP baselines, we use Gemma-3-1B-PT \citep{team2025gemma3}, a 1B-parameter model from the Gemma-3 family that matches VaultGemma-1B in size, isolating the effect of DP from model capacity; and Gemma-2-2B \citep{team2024gemma2}, a 2B-parameter model from the same Gemma-2 family as VaultGemma-1B, isolating the effect of DP from training family differences while allowing comparison with a larger model. We compare these against VaultGemma-1B \citep{sinha2025vaultgemma}, a 1B-parameter model pretrained with DP-SGD on the Gemma-2 architecture with a privacy budget of $\varepsilon=2$, where $\varepsilon$ bounds the influence of any individual training example on the model. To our knowledge, VaultGemma-1B is currently the only publicly available LLM pretrained with DP-SGD, making it the sole candidate for this evaluation. All three are pretrained base models with no instruction tuning.

\subsection{Implementation Details}

All experiments are conducted on Tesla V100 GPUs (32GB) using PyTorch \citep{paszke2019pytorch} and the HuggingFace Transformers library \citep{wolf2020transformers}. For sentence scoring benchmarks, inputs are not truncated, and no generation is performed. For text completion, we generate up to 50 tokens per prompt. For BBQ, we generate up to 5 tokens and extract the first token matching A, B, or C as the model’s answer; if no such token appears, the item is recorded as a parsing failure and excluded from accuracy and SS computation. For tabular classification, model outputs are parsed to extract a binary prediction; unparseable responses are similarly excluded.

\begin{table}[t]
\centering
\small
\caption{Models evaluated. All are pretrained base models with no instruction tuning.}
\begin{tabular}{lccc}
\toprule
\textbf{Model} & \textbf{Size} & \textbf{DP} & \textbf{Training family} \\
\midrule
VaultGemma-1B & 1B & Yes ($\varepsilon=2$) & Gemma-2 \\
Gemma-3-1B-PT & 1B & No & Gemma-3 \\
Gemma-2-2B & 2B & No & Gemma-2 \\
\bottomrule
\end{tabular}
\label{tab:models}
\end{table}


\subsection{Metrics}

We use different metrics for each evaluation paradigm, described below.

\subsubsection{Sentence Scoring Metrics}

For each example $i$, we compute the normalized log-probabilities of the stereotypical, anti-stereotypical, and unrelated candidate sentences, denoted as $\ell_i^+$, $\ell_i^-$, and $\ell_i^u$, respectively. These scores quantify the model's relative preference for each candidate given the input context.

where normalized log-probability for sentence $s$ is: 
\begin{equation}
\small
\ell(s) = \frac{1}{|s|} \sum_{t=1}^{|s|} \log P(w_t \mid w_{<t})
\end{equation}
and $|s|$ is the number of tokens in $s$.
Then we evaluate the model using three metrics following \citet{nadeem2021stereoset}.

\paragraph{Language Model Score (LMS)} evaluates the capacity of the model to assign a higher likelihood to a meaningful sentence (stereotype or anti-stereotype option) over an unrelated sentence. 
\begin{equation}
\small
\begin{aligned}
\textbf{LMS} &= \frac{\sum_{i=1}^{N} \mathbf{1}[\max(\ell_i^+, \ell_i^-) > \ell_i^u]}{N} \times 100 \\
\end{aligned}
\end{equation}

\paragraph{Stereotype Score (SS)} measures the stereotype preference between the two meaningful options, where 50 indicates a neutral model. 
\begin{equation}
\label{eq:ss}
\small
\begin{aligned}
\textbf{SS} &= \frac{\sum_{i=1}^{N} \mathbf{1}[\ell_i^+ > \ell_i^-] \cdot \mathbf{1}[\max(\ell_i^+, \ell_i^-) > \ell_i^u]}{\sum_{i=1}^{N} \mathbf{1}[\max(\ell_i^+, \ell_i^-) > \ell_i^u]} \times 100 \\
\end{aligned}
\end{equation}

\paragraph{Idealized Context Association Test (ICAT)} combines both by penalising models whose SS deviates from 50, rewarding models that are both fluent and unbiased where 100 corresponds to the ideal score.
\begin{equation}
\small
\begin{aligned}
\textbf{ICAT} &= \textbf{LMS}\times \frac{\min(\textbf{SS},\ 100 - \textbf{SS})}{50}
\end{aligned}
\end{equation}

\subsubsection{Text Completion Metrics}

\paragraph{Regard} We use a pre-trained BERT model \cite{sheng2019woman} that was fine-tuned on human-annotated data to measure regard towards demographics indicated by language as opposed to overall sentiment. The model predicts four labels (positive, neutral, negative, other) for input text. We report the argmax of the regarded labels for each generated completion, excluding the other label and retaining only positive,   
  neutral, and negative. We focus on measuring regard for gender and race demographics since the model was trained on these two demographics only.

\paragraph{Sentiment} We use the Valence Aware Dictionary and Sentiment Reasoner (VADER) \cite{hutto2014vader}, which computes the sentiment score of a text by first assigning word-level valence-based lexicon scores and then combining them with rules that account for contextual information. For each text, VADER produces a score in the range $[-1, 1]$, where $-1$ represents negative sentiment and $1$ represents positive sentiment. Using texts with known sentiment labels, we determine thresholds of $\geq 0.5$ and $\leq -0.5$ to classify texts as conveying positive and negative sentiments, respectively.

\paragraph{Toxicity} We evaluate the toxicity of generated text using Detoxify \cite{detoxify}, a pre-trained toxicity classifier based on Transformer architectures and fine-tuned on toxic comment datasets. Detoxify assigns a score to each input text for several different categories of toxicity, including toxicity, severe toxicity, obscene, identity attack, insult, and threat. We classify a text as toxic if the maximum score across all categories exceeds $0.5$.

\paragraph{Bias Gap} 
For each metric, we compute the signed gaps $\Delta_{\text{gender}} = p_{\text{male}} - p_{\text{female}}$ and $\Delta_{\text{race}} = p_{\text{white}} - p_{\text{black}}$ for each label (positive, neutral, and negative). As a summary fairness measure, we report the mean absolute gap:
\begin{equation}
\overline{|\Delta|} = \frac{|\Delta_{\text{pos}}| + |\Delta_{\text{neu}}| + |\Delta_{\text{neg}}|}{3}
\end{equation}
A smaller $\overline{|\Delta|}$ indicates more equitable behavior across groups \cite{dwork2012fairness, dhamala2021bold}.

\subsubsection{Tabular Classification Metrics}
For the task of evaluating tabular data, we reformulate the problem as a binary classification task and measure performance in terms of both accuracy and group fairness with respect to a protected attribute (race or sex, depending on the dataset).

\paragraph{Accuracy} is the ratio of correctly predicted labels to the total number of model outputs that were parseable.

\begin{equation}
\label{eq:acc}
\text{\textbf{Acc}} = \frac{\sum_{i=1}^{N} \mathbf{1}[\hat{y}_i = y_i]}{N}
\end{equation}

\paragraph{Demographic Parity Difference (DPD)} is defined as the absolute difference of the positive prediction rates of the two groups of interest (in this case racial groups) \cite{dwork2012fairness}.

\begin{equation}
\small
\text{\textbf{DPD}} = |P(\hat{y}=1 \mid g=\text{White}) - P(\hat{y}=1 \mid g=\text{Black})|
\end{equation}

\paragraph{Equalized Odds Difference (EoD)} is defined as the maximum difference in true positive rates (TPR) and false positive rates (FPR) across groups \cite{hardt2016equality}.

\begin{equation}
\small
\text{\textbf{EoD}} = \max\left(|\text{TPR}_W - \text{TPR}_B|,\ |\text{FPR}_W - \text{FPR}_B|\right)
\end{equation}

\paragraph{Equal Opportunity Difference (EoOpp)} is defined as the absolute difference in true positive rates between groups \cite{hardt2016equality}.

\begin{equation}
\small
\text{\textbf{EoOpp}} = |\text{TPR}_W - \text{TPR}_B|
\end{equation}

For all fairness metrics, lower values indicate more equitable model behavior. The demographic parity and equalized odds metrics were calculated using the \texttt{fairlearn} library \cite{bird2020fairlearn}.

\subsubsection{Question Answering Metrics}

We evaluate models on BBQ using two metrics. Each item consists of a context with three answer options: a stereotyped answer, an anti-stereotyped answer, and \textit{``Can't be determined''}. We report \textbf{Accuracy} on disambiguated contexts (\cref{eq:acc}) and \textbf{Stereotype Score (SS)} on ambiguous contexts (\cref{eq:ss}). The score is computed using $n_{\text{target}}$, the number of times the model selects the stereotyped answer, and $n_{\text{anti-target}}$, the number of anti-stereotyped selections, both excluding \textit{``Can't be determined''} responses. A score of 50 indicates an unbiased model.

\section{Experimental Results}

\subsection{Sentence Scoring Results}
\begin{table}[h!]
    \centering
    \small
        \caption{Bias results on Sentence Scoring datasets. \textbf{Bold} = best ICAT per dataset.}
    \resizebox{\columnwidth}{!}{
    \begin{tabular}{ccccc}
    \toprule
        \textbf{Dataset} & \textbf{Model} & \textbf{LMS} $\uparrow$ & \textbf{SS}$\rightarrow 50$ & \textbf{ICAT} $\uparrow$  \\
    \midrule
         \multirow{3}{*}{\textbf{StereoSet-Intra}} & VaultGemma-1B & 96.98 & 62.62 & \textbf{72.49}\\
                                         & Gemma-3-1B-PT & 97.17 & 64.36 & 69.26 \\
                                         & Gemma-2-2B & 97.23 & 66.57 & 65.01 \\
         \midrule
         \multirow{3}{*}{\textbf{StereoSet-Inter}} & VaultGemma-1B & 86.86 & 51.95 & \textbf{83.47} \\
                                         & Gemma-3-1B-PT & 88.46 & 54.05 & 81.30 \\
                                         & Gemma-2-2B & 90.34 & 57.56 & 76.68 \\
         \midrule
         \multirow{3}{*}{\textbf{WinoBias Type 1}} & VaultGemma-1B & 100.00 & 50.25 & \textbf{99.49} \\
                                         & Gemma-3-1B-PT & 100.00 & 58.84 & 82.32 \\
                                         & Gemma-2-2B & 100.00 & 62.12 & 75.76 \\
         \midrule
         \multirow{3}{*}{\textbf{WinoBias Type 2}} & VaultGemma-1B & 100.00 & 61.62 & 76.77\\
                                         & Gemma-3-1B-PT & 100.00 & 61.36 & \textbf{77.27}\\
                                         & Gemma-2-2B & 100.00 & 61.87 & 76.26 \\                               
         \midrule
         \multirow{3}{*}{\textbf{CrowS-Pairs}} & VaultGemma-1B & 97.68 & 58.04 & \textbf{81.96}\\
                                         & Gemma-3-1B-PT & 97.75 & 62.96 & 72.41 \\
                                         & Gemma-2-2B & 98.01 & 64.34 & 69.89 \\
     \bottomrule
    \end{tabular}
    }
    \label{tab:Likelihood-Based}
\end{table}

We evaluate all models on Sentence Scoring benchmarks, as shown in~\cref{tab:Likelihood-Based}. Overall, the VaultGemma-1B model generally scores lower in the stereotype score (SS) while scoring higher on the ICAT than non-DP models, indicating a weaker preference for the stereotypical option. On the other hand, Gemma-2-2B scores marginally higher in language model scores (LMS) but exhibits higher stereotype preference as well as lower ICAT scores.

On StereoSet (both intra- and inter-sentence), VaultGemma-1B achieves the best ICAT and lowest SS among all models. A similar pattern is observed on CrowS-Pairs, where the DP model consistently outperforms non-DP models in terms of reduced stereotype preference. On WinoBias Type~1, VaultGemma-1B achieves near-neutral SS values, while non-DP models exhibit stronger bias, as reflected in higher SS and lower ICAT.

In contrast, all models perform similarly on WinoBias Type~2, where SS and ICAT values are comparable across models. This suggests that performance differences are minimal when the task can be solved using syntactic cues alone, without requiring reliance on stereotypical associations. Overall, these results suggest that privacy-preserving training is associated with reduced stereotype preference in Sentence Scoring evaluations. Per-domain breakdowns are provided in~\cref{tab:combined_results} in the Appendix.

\subsection{Text Completion Results}

\begin{table*}[t]
\centering
\small
\caption{Regard and sentiment analysis results across BOLD and HolisticBias. \textbf{Bold} = lowest $\overline{|\Delta|}$ per group and metric.}
\begin{tabular}{lllcccc|cccc}
\toprule
\multirow{2}{*}{\textbf{Dataset}} & \multirow{2}{*}{\textbf{Metric}} & \multirow{2}{*}{\textbf{Model}}
& \multicolumn{4}{c}{\textbf{Gender (male$-$female)}}
& \multicolumn{4}{c}{\textbf{Race (white$-$black)}} \\
\cmidrule(lr){4-7} \cmidrule(lr){8-11}
& & 
& $\Delta_{\text{pos}}$ $\downarrow$ 
& $\Delta_{\text{neu}}$ $\downarrow$ 
& $\Delta_{\text{neg}}$ $\downarrow$ 
& $\overline{|\Delta|}$ $\downarrow$
& $\Delta_{\text{pos}}$ $\downarrow$ 
& $\Delta_{\text{neu}}$ $\downarrow$ 
& $\Delta_{\text{neg}}$ $\downarrow$ 
& $\overline{|\Delta|}$ $\downarrow$ \\
\midrule

\multirow{6}{*}{\textbf{BOLD}}
& \multirow{3}{*}{Regard}
& VaultGemma-1B & $+0.52$ & $-1.12$ & $+0.61$ & $\mathbf{0.75}$ & $+1.51$ & $-0.43$ & $-1.08$ & $\mathbf{1.01}$ \\
& & Gemma-3-1B-PT  & $-8.05$ & $+6.40$ & $+1.64$ & $5.36$ & $+0.65$ & $+2.05$ & $-2.70$ & $1.80$ \\
& & Gemma-2-2B  & $-5.97$ & $+4.07$ & $+1.90$ & $3.98$ & $-0.65$ & $+3.51$ & $-2.86$ & $2.34$ \\
\cmidrule(lr){2-11}

& \multirow{3}{*}{Sentiment}
& VaultGemma-1B & $-3.89$ & $+2.16$ & $+1.73$ & $2.60$ & $-4.64$ & $+2.43$ & $+2.21$ & $3.09$ \\
& & Gemma-3-1B-PT  & $-9.69$ & $+6.57$ & $+3.11$ & $6.46$ & $-2.64$ & $+4.21$ & $-1.56$ & $\mathbf{2.80}$ \\
& & Gemma-2-2B  & $-3.20$ & $-0.17$ & $+3.37$ & $\mathbf{2.25}$ & $-5.72$ & $+5.02$ & $+0.70$ & $3.81$ \\

\midrule

\multirow{6}{*}{\textbf{HolisticBias}}
& \multirow{3}{*}{Regard}
& VaultGemma-1B & $-3.52$ & $+2.28$ & $+1.24$ & $2.35$ & $-1.80$ & $+5.05$ & $-3.25$ & $\mathbf{3.37}$ \\
& & Gemma-3-1B-PT  & $-1.24$ & $-0.12$ & $+1.36$ & $\mathbf{0.90}$ & $-2.95$ & $+6.42$ & $-3.47$ & $4.28$ \\
& & Gemma-2-2B  & $-1.53$ & $+2.94$ & $-1.41$ & $1.96$ & $-3.61$ & $+7.43$ & $-3.81$ & $4.95$ \\
\cmidrule(lr){2-11}

& \multirow{3}{*}{Sentiment}
& VaultGemma-1B & $-2.36$ & $+3.23$ & $-0.87$ & $2.15$ & $+2.59$ & $-1.35$ & $-1.24$ & $1.73$ \\
& & Gemma-3-1B-PT  & $+0.61$ & $-0.40$ & $-0.20$ & $\mathbf{0.40}$ & $+0.47$ & $+2.05$ & $-2.52$ & $\mathbf{1.68}$ \\
& & Gemma-2-2B  & $-0.40$ & $+0.63$ & $-0.23$ & $0.42$ & $+1.94$ & $+1.24$ & $-3.18$ & $2.12$ \\

\bottomrule
\end{tabular}
\label{tab:regard/sentiment}
\end{table*}

\begin{table*}[t]
\centering
\small
\caption{Toxicity analysis across BOLD and HolisticBias datasets (\%). \textbf{Bold} = smallest bias gap per demographic.}
\begin{tabular}{llccc|ccc}
\toprule
\multirow{2}{*}{\textbf{Dataset}} & \multirow{2}{*}{\textbf{Model}}
& \multicolumn{3}{c}{\textbf{Gender (male vs.\ female)}}
& \multicolumn{3}{c}{\textbf{Race (white vs.\ black)}} \\
\cmidrule(lr){3-5} \cmidrule(lr){6-8}
& & Male $\downarrow$ & Female $\downarrow$ & $|\Delta_{\text{gender}}|$ $\downarrow$
& Black $\downarrow$ & White $\downarrow$ & $|\Delta_{\text{race}}|$ $\downarrow$ \\
\midrule

\multirow{3}{*}{\textbf{BOLD}}
& VaultGemma-1B & $0.00$ & $0.09$ & $0.09$ & $0.38$ & $0.11$ & $0.27$ \\
& Gemma-3-1B-PT    & $0.09$ & $0.00$ & $0.09$ & $0.49$ & $0.00$ & $0.49$ \\
& Gemma-2-2B    & $0.00$ & $0.00$ & $\mathbf{0.00}$ & $0.27$ & $0.32$ & $\mathbf{0.05}$ \\

\midrule

\multirow{3}{*}{\textbf{HolisticBias}}
& VaultGemma-1B & $3.52$ & $3.26$ & $0.26$ & $3.00$ & $2.21$ & $\mathbf{0.79}$ \\
& Gemma-3-1B-PT    & $5.08$ & $4.99$ & $\mathbf{0.09}$ & $5.30$ & $3.20$ & $2.10$ \\
& Gemma-2-2B    & $6.14$ & $5.48$ & $0.66$ & $5.61$ & $3.56$ & $2.05$ \\

\bottomrule
\end{tabular}
\label{tab:toxicity}
\end{table*}

We report bias evaluation results on Text Completion benchmarks in Tables~\ref{tab:regard/sentiment} and~\ref{tab:toxicity}. Compared to Sentence Scoring evaluations, the results exhibit a more mixed pattern across models and metrics.

For the regard metric, VaultGemma-1B achieves smaller bias gaps than the non-DP models for both gender and racial groups on the BOLD dataset. On HolisticBias, VaultGemma-1B achieves the smallest bias gap for race, while Gemma-3-1B-PT marginally achieves the smallest bias gap for gender. Based on this metric, the DP model remains competitive, achieving the best performance in 3 out of 4 cases.

In terms of sentiment, the results show a different pattern compared to Sentence Scoring metrics. On BOLD, Gemma-2-2B obtains the smallest gender bias gap, while on HolisticBias, Gemma-3-1B-PT achieves the lowest gap for race. For gender on HolisticBias, both Gemma-3-1B-PT and Gemma-2-2B show similarly small gaps. On the other hand, VaultGemma-1B achieves gaps comparable to Gemma-3-1B-PT only in the race setting.

For toxicity, all models exhibit very low toxicity levels on BOLD, resulting in minimal differences across models, limiting interpretability of the results. Nevertheless, on HolisticBias clearer distinctions emerge: VaultGemma-1B achieves the lowest toxicity levels for both gender and race, as well as the smallest race-based toxicity gap. 

Overall, the DP model performs well on regard and toxicity, but results on text completion benchmarks are mixed across models and metrics.

\subsection{Tabular Classification Results}

\begin{table*}[h!]
\centering
\caption{Accuracy and fairness metrics under four in-context learning (ICL) strategies evaluated on the Adult Census Income dataset with race as the protected attribute. Higher accuracy is better, while lower values indicate improved fairness for DPD, EoD, and EoOpp. No bolding: no consistent model advantage across settings.}
\resizebox{\textwidth}{!}{%
\begin{tabular}{ll cccc|cccc|cccc}
\toprule
& & \multicolumn{4}{c}{VaultGemma-1B} & \multicolumn{4}{c}{Gemma-3-1B-PT} & \multicolumn{4}{c}{Gemma-2-2B} \\
\cmidrule(lr){3-6} \cmidrule(lr){7-10} \cmidrule(lr){11-14}
Selection & $k$ & Acc & DPD & EoD & EoOpp
& Acc & DPD & EoD & EoOpp
& Acc & DPD & EoD & EoOpp \\
\midrule
\multirow{3}{*}{Random} 
& 0 & 53.80 & 0.04 & 0.05 & 0.05 & 50.00 & 0.00 & 0.00 & 0.00 & 50.00 & 0.00 & 0.00 & 0.00 \\
& 4 & 51.67 & 0.02 & 0.04 & 0.02 & 55.70 & 0.05 & 0.09 & 0.09 & 55.50 & 0.06 & 0.10 & 0.10 \\
& 8 & 49.07 & 0.02 & 0.04 & 0.04 & 51.67 & 0.01 & 0.03 & 0.02 & 57.40 & 0.07 & 0.12 & 0.12 \\
\midrule
\multirow{3}{*}{Balanced} 
& 0 & 53.80 & 0.04 & 0.05 & 0.05 & 50.00 & 0.00 & 0.00 & 0.00 & 50.00 & 0.00 & 0.00 & 0.00 \\
& 4 & 54.30 & 0.03 & 0.05 & 0.04 & 66.30 & 0.08 & 0.12 & 0.07 & 64.73 & 0.07 & 0.11 & 0.11 \\
& 8 & 58.87 & 0.03 & 0.05 & 0.03 & 60.10 & 0.07 & 0.09 & 0.09 & 57.73 & 0.01 & 0.03 & 0.02 \\
\midrule
\multirow{3}{*}{Instruction} 
& 0 & 50.00 & 0.00 & 0.00 & 0.00 & 50.00 & 0.00 & 0.00 & 0.00 & 50.00 & 0.00 & 0.00 & 0.00 \\
& 4 & 51.43 & 0.00 & 0.02 & 0.02 & 55.53 & 0.08 & 0.11 & 0.11 & 53.10 & 0.05 & 0.08 & 0.08 \\
& 8 & 49.23 & 0.01 & 0.02 & 0.02 & 52.63 & 0.02 & 0.03 & 0.03 & 55.03 & 0.06 & 0.10 & 0.10 \\
\midrule
\multirow{3}{*}{Removal} 
& 0 & 49.40 & 0.04 & 0.06 & 0.06 & 50.00 & 0.00 & 0.00 & 0.00 & 50.00 & 0.00 & 0.00 & 0.00 \\
& 4 & 51.60 & 0.01 & 0.03 & 0.02 & 55.43 & 0.03 & 0.05 & 0.05 & 57.37 & 0.01 & 0.03 & 0.03 \\
& 8 & 50.10 & 0.01 & 0.02 & 0.02 & 52.33 & 0.00 & 0.01 & 0.01 & 57.70 & 0.04 & 0.08 & 0.08 \\
\bottomrule
\end{tabular}%
}
\label{tab:fairness_results_adult}
\end{table*}

Table~\ref{tab:fairness_results_adult} presents accuracy and fairness metrics on the Adult Census Income dataset, where models predict whether an individual earns more than \$50K annually, with race as the protected attribute. At zero shots, all models perform at chance level with negligible fairness gaps, consistent with an inability to perform the task without demonstrations. Providing in-context examples substantially improves accuracy but consistently introduces racial bias across all models and selection strategies, suggesting that the demonstrations themselves are a primary driver of model behavior in this setting.

Across selection strategies, no consistent advantage emerges for VaultGemma-1B over its non-DP counterparts. The fairness metrics vary more as a function of the demonstration selection strategy than of the model itself, making it difficult to attribute differences in bias to DP training.

On COMPAS (Table~\ref{tab:fairness_results_compas}), all models achieve accuracy near the 50\% random baseline across all strategies and shot counts. Inspection of predictions reveals that models default to predicting a single class for all samples, making both accuracy and fairness metrics uninformative. A similar pattern holds on German Credit (Table~\ref{tab:fairness_results_german}), where accuracy remains close to chance and no consistent model differences emerge. These results suggest that pretrained base models may not reliably extract predictive signal from structured tabular demonstrations via ICL. We discuss the implications further in Section~\ref{sec:disc}.

\subsection{Question Answering Results}

\begin{table*}[t]
\centering
\caption{BBQ results across demographic domains. Accuracy is near the random baseline (1/3); SS values closer to 50 indicate less bias. \textbf{Bold} = best Acc per category.}
\begin{tabular}{lcc|cc|cc|cc}
\toprule
\multirow{2}{*}{\textbf{Model}} & \multicolumn{2}{c}{\textbf{Gender}}
& \multicolumn{2}{c}{\textbf{Nationality}}
& \multicolumn{2}{c}{\textbf{Race}}
& \multicolumn{2}{c}{\textbf{Religion}} \\
\cmidrule(lr){2-3} \cmidrule(lr){4-5} \cmidrule(lr){6-7} \cmidrule(lr){8-9}
& Acc $\uparrow$ & SS $\rightarrow 50$
& Acc $\uparrow$ & SS $\rightarrow 50$
& Acc $\uparrow$ & SS $\rightarrow 50$
& Acc $\uparrow$ & SS $\rightarrow 50$ \\
\midrule

VaultGemma-1B & 0.36 & 51.45 & 0.31 & 49.46 & 0.32 & 48.95 & 0.32 & 50.25 \\
Gemma-3-1B-PT & 0.34 & 49.11 & 0.37 & 51.35 & 0.39 & 51.83 & 0.41 & 51.53 \\
Gemma-2-2B    & \textbf{0.50} & 51.60 & \textbf{0.45} & 49.87 & \textbf{0.49} & 46.12 & \textbf{0.49} & 49.53 \\

\bottomrule
\end{tabular}
\label{tab:bbq}
\end{table*}

Table~\ref{tab:bbq} presents results on the BBQ benchmark across four demographic categories. All models produce parseable outputs with near-zero failure rates. Accuracy on disambiguated contexts ranges from 0.33 to 0.48, close to the random baseline of 0.33 for a three-choice setting, indicating that pretrained base models struggle to use contextual information to resolve the correct answer. Stereotype Scores (SS) remain consistently close to 50 across all models and categories, suggesting no systematic preference for stereotyped or anti-stereotyped responses. As a result, these evaluations do not reliably measure model-intrinsic bias, but rather reflect the inability of base models to follow the task format.

\section{Discussion}
\label{sec:disc}

\paragraph{Sentence scoring}
Across Sentence Scoring benchmarks, DP consistently reduces stereotype-driven preferences, as reflected by lower SS and higher ICAT scores. One possible explanation is that DP-SGD limits the model’s ability to memorize high-frequency co-occurrence patterns between demographic groups and stereotypical attributes. By introducing noise and clipping gradients during training \citep{abadi2016deep}, DP weakens the influence of individual training examples, effectively smoothing the learned probability distribution.

As a result, the model becomes less confident in assigning higher likelihood to stereotypical completions, leading to more balanced preferences between stereotypical and anti-stereotypical options. This effect is particularly pronounced in Sentence Scoring evaluations, where bias is measured directly through relative log-probabilities. In this setting, even small shifts in probability mass can lead to substantial changes in model preference, making these metrics especially sensitive to the effects of DP.

This interpretation is further supported by the distinction between WinoBias Type~1 and Type~2. In Type~1, assigning higher likelihood to one sentence over another requires reliance on world knowledge and learned gender-occupation associations, and we observe that DP substantially reduces bias in this setting. In contrast, in Type~2, where the difference between candidate sentences is determined by syntactic cues (e.g., grammatical structure), all models behave similarly regardless of DP. This contrast suggests that DP primarily weakens reliance on learned stereotypical associations, while having little effect when decisions can be made based on syntax alone.

\paragraph{Text completion}

The effects of DP in text completion are less consistent across metrics such as regard, sentiment, and toxicity. While reductions in demographic disparities are observed in some cases, the results depend heavily on the dataset and evaluation metric. This discrepancy can be explained by the fact that open-ended generation involves additional factors beyond token-level probabilities, including decoding strategies and surface-level lexical choices.

Moreover, metrics such as sentiment may capture overall tone rather than deeper stereotypical associations, while toxicity scores are often uniformly low, limiting their ability to distinguish between models. As a result, improvements observed in Sentence Scoring evaluations do not necessarily translate into consistent changes in generated outputs. These findings suggest that bias reduction at the logit level does not reliably transfer to Text Completion, highlighting a gap between logit-level and output-level bias.

\paragraph{Tabular classification}

In the tabular classification setting, no consistent advantage emerges for the DP model. Overall performance remains close to chance level, with accuracy around 50\%, indicating that pretrained base models struggle to reliably perform this task. Instead, both accuracy and fairness metrics are strongly influenced by the choice of in-context learning (ICL) strategy. Providing demonstrations often introduces bias across all models, regardless of whether they are trained with DP.

One possible explanation is that pretrained models are not well-aligned for structured question answering tasks on tabular datasets, and therefore rely heavily on superficial patterns or positional biases rather than meaningful reasoning. As a result, predictions tend to be unstable and sensitive to prompt design, which can dominate any bias-related effects introduced during pretraining. This suggests that in-context evaluations on tabular datasets may not be well-suited for assessing model-intrinsic bias in pretrained base models, as they conflate limitations in task performance with fairness outcomes.

\paragraph{Question Answering}
On BBQ, all models exhibit SS values near 50 and accuracy close to the random baseline, making it difficult to draw conclusions about the effect of differential privacy on bias. The near-random accuracy indicates that pretrained base models do not reliably follow the multiple-choice (A/B/C) format, instead producing outputs that are effectively arbitrary. This undermines the validity of the SS metric, as random selection naturally yields scores close to 50 regardless of any underlying bias. These results suggest that BBQ, in its generation-based formulation, is not well-suited for evaluating pretrained base models, which lack instruction tuning and therefore struggle to adhere to the task format.

\paragraph{Implications for evaluating bias in LLMs.}

Taken together, our results show that the impact of DP on social bias depends strongly on the evaluation setting. While DP reduces stereotype-driven preferences in Sentence Scoring evaluations, its effects are less consistent in Text Completion and largely absent in Tabular Classification settings. This indicates that bias is not a single, uniform property of a model, but rather depends on how it is measured. 

More broadly, these findings reveal a disconnect between logit-level and output-level bias. Reducing bias at the logit level by weakening memorized associations does not necessarily guarantee fairer outputs across other evaluation paradigms. This highlights the importance of using multiple complementary evaluation paradigms when assessing fairness in LLMs, as relying on a single benchmark may provide an incomplete or misleading picture of model behavior.

\section{Limitations}
\label{sec:limitations}
Our analysis is constrained by the availability of a single publicly released DP-pretrained LLM. While this limits comparisons across architectures and scales, it also reflects the current state of DP training for large language models and provides a controlled setting to isolate the effects of DP at the pretraining stage.

We evaluate pretrained base models, as DP has only been applied at the pretraining stage in currently available LLMs. This setting allows us to directly study how DP influences learned representations before any post-training alignment. Extending this analysis to instruction-tuned or fine-tuned DP models is an important direction for future work.

\section{Conclusion}

In this work, we presented a systematic evaluation of the effect of differential privacy on social bias in pretrained LLMs across four complementary paradigms. Our results show that DP reduces stereotypical bias in Sentence Scoring evaluations, but these improvements do not consistently transfer to other evaluation paradigms such as text generation, tabular classification, and question answering, where outcomes are often influenced by task limitations and evaluation design. These findings highlight a disconnect between logit-level and output-level bias, suggesting that bias is not a single property but depends on how it is measured. Overall, our work underscores the importance of multi-paradigm evaluation and suggests that privacy-preserving training alone is insufficient to ensure equitable model outputs.

\FloatBarrier
\section*{Acknowledgements}
This work is supported in part by the National Science Foundation under awards 1910284, 1946391, and 2147375.

\bibliography{latex/references}

\clearpage
\appendix









\section{Sentence Scoring: Per-Domain Results}

Table~\ref{tab:combined_results} presents per-domain breakdowns of LMS, SS, and ICAT across all Sentence Scoring benchmarks. The DP advantage observed in overall scores holds across most domains, but with notable variation. The most pronounced gains for VaultGemma-1B appear in the religion and race domains, where ICAT differences relative to non-DP models are largest. In contrast, the profession domain on StereoSet-Intra is the one case where Gemma-3-1B-PT marginally outperforms the DP model. On CrowS-Pairs, the sexual orientation category is the only domain where VaultGemma-1B does not achieve the best ICAT, with Gemma-2-2B scoring slightly higher. These domain-level differences suggest that the bias-reducing effect of DP is not uniform across all social categories.

\begin{table*}
  \centering
  \caption{Bias evaluation results across StereoSet, CrowS-Pairs, and WinoBias. Higher LMS and ICAT indicate better performance, while SS closer to 50 indicates less bias. \textbf{Bold} indicates the highest ICAT (best fairness) per row.}
  \small
  \setlength{\tabcolsep}{3pt}

  \begin{tabular}{llccc|ccc|ccc}
  \toprule
  \multirow{2}{*}{\textbf{Dataset}} & \multirow{2}{*}{\textbf{Domain}}
  & \multicolumn{3}{c|}{VaultGemma-1B}
  & \multicolumn{3}{c|}{Gemma-3-1B-PT}
  & \multicolumn{3}{c}{Gemma-2-2B} \\

  \cmidrule(lr){3-11}
  & & LMS$\uparrow$ & SS$\rightarrow50$ & ICAT$\uparrow$
    & LMS$\uparrow$ & SS$\rightarrow50$ & ICAT$\uparrow$
    & LMS$\uparrow$ & SS$\rightarrow50$ & ICAT$\uparrow$ \\
  \midrule

  \multirow{5}{*}{\textbf{Stereoset-Intra}}
  & Gender     & 98.82 & 69.44 & \textbf{60.39} & 97.65 & 69.88 & 58.82 & 97.65 & 75.10 & 48.63 \\
  & Profession & 96.54 & 65.60 & 66.42 & 96.54 & 65.22 & \textbf{67.16} & 96.67 & 68.20 & 61.48 \\
  & Race       & 96.88 & 59.01 & \textbf{79.42} & 97.61 & 61.87 & 74.43 & 97.51 & 63.33 & 71.52 \\
  & Religion   & 97.47 & 57.14 & \textbf{83.54} & 97.47 & 68.83 & 60.76 & 98.73 & 62.82 & 73.42 \\
  \cmidrule{2-11}
  & Overall & 96.98 & 62.62 & \textbf{72.49} & 97.17 & 64.36 & 69.26 & 97.23 & 66.57 & 65.01 \\
  \midrule

  \multirow{5}{*}{\textbf{Stereoset-Inter}}
  & Gender     & 86.78 & 60.00 & \textbf{69.42} & 86.36 & 62.20 & 65.29 & 88.43 & 64.02 & 63.64 \\
  & Profession & 87.06 & 54.72 & \textbf{78.84} & 89.24 & 55.83 & \textbf{78.84} & 91.66 & 59.89 & 73.52 \\
  & Race       & 86.58 & 48.05 & 83.20 & 88.11 & 50.35 & \textbf{87.50} & 89.45 & 53.95 & 82.38 \\
  & Religion   & 88.46 & 46.38 & \textbf{82.05} & 91.03 & 56.34 & 79.49 & 93.59 & 57.53 & 79.49 \\
  \cmidrule{2-11}
  & Overall & 86.86 & 51.95 & \textbf{83.47} & 88.46 & 54.05 & 81.30 & 90.34 & 57.56 & 76.68 \\
  \midrule

  \multirow{10}{*}{\textbf{CrowS-Pairs}}
  & Gender             & 98.85 & 55.21 & \textbf{88.55} & 97.71 & 60.94 & 76.34 & 98.47 & 64.34 & 70.23 \\
  & Race               & 96.12 & 55.44 & \textbf{85.66} & 96.51 & 61.85 & 73.64 & 97.09 & 60.68 & 76.36 \\
  & Nationality        & 98.74 & 46.50 & \textbf{91.82} & 98.74 & 46.50 & \textbf{91.82} & 98.11 & 56.41 & 85.53 \\
  & Religion           & 97.14 & 62.75 & \textbf{72.38} & 97.14 & 62.75 & \textbf{72.38} & 96.19 & 68.32 & 60.95 \\
  & Age                & 98.85 & 61.63 & \textbf{75.86} & 98.85 & 62.79 & 73.56 & 98.85 & 63.95 & 71.26 \\
  & Disability         & 98.33 & 64.41 & 70.00 & 98.33 & 61.02 & \textbf{76.67} & 98.33 & 66.10 & 66.67 \\
  & Phys.\ appearance  & 98.41 & 62.90 & \textbf{73.02} & 98.41 & 74.19 & 50.79 & 100.00 & 73.02 & 53.97 \\
  & Sexual orientation & 98.81 & 78.31 & 42.86 & 98.81 & 84.34 & 30.95 & 97.62 & 75.61 & \textbf{47.62} \\
  & Socioeconomic      & 98.26 & 62.13 & \textbf{74.42} & 99.42 & 70.76 & 58.14 & 100.00 & 70.93 & 58.14 \\
  \cmidrule{2-11}
  & Overall   & 97.68 & 58.04 & \textbf{81.96} & 97.75 & 62.96 & 72.41 & 98.01 & 64.34 & 69.89 \\
  \midrule

  \multirow{2}{*}{\textbf{WinoBias Type 1}}
  & Gender & 100.00 & 50.25 & \textbf{99.49} & 100.00 & 58.84 & 82.32 & 100.00 & 62.12 & 75.76 \\
  \cmidrule{2-11}
  & Overall & 100.00 & 50.25 & \textbf{99.49} & 100.00 & 58.84 & 82.32 & 100.00 & 62.12 & 75.76 \\
  \midrule

  \multirow{2}{*}{\textbf{WinoBias Type 2}}
  & Gender & 100.00 & 61.62 & 76.77 & 100.00 & 61.36 & \textbf{77.27} & 100.00 & 61.87 & 76.26 \\
  \cmidrule{2-11}
  & Overall & 100.00 & 61.62 & 76.77 & 100.00 & 61.36 & \textbf{77.27} & 100.00 & 61.87 & 76.26 \\
  \bottomrule
  \end{tabular}

  \label{tab:combined_results}
\end{table*}

\section{Tabular Data Serialization Example}
\label{fig:serialization}

\begin{figure}[h]
\centering
\small
\fbox{%
\begin{minipage}{0.9\columnwidth}
\textbf{[Demonstration]}\\
\textit{Input:} age is 45, workclass is Private, education is Bachelors, marital-status is Married-civ-spouse, occupation is Exec-managerial, race is White, sex is Male, hours-per-week is 50. Does this person earn more than 50K annually?\\
\textit{Output:} Yes\\[4pt]
\textbf{[Test]}\\
\textit{Input:} age is 28, workclass is Self-emp-not-inc, education is HS-grad, marital-status is Never-married, occupation is Craft-repair, race is Black, sex is Male, hours-per-week is 40. Does this person earn more than 50K annually?\\
\textit{Output:}
\end{minipage}%
}
\caption{Example of tabular data serialization for Adult Census Income with one demonstration ($k=1$). Each feature is converted to a ``\textit{key} is \textit{value}'' phrase, followed by a dataset-specific binary question. The same format is used for COMPAS and German Credit.}
\end{figure}

\section{Tabular Classification: Full Results}

Tables~\ref{tab:fairness_results_compas} and~\ref{tab:fairness_results_german} present full accuracy and fairness results on COMPAS and German Credit. Consistent with the Adult Census Income results in the main paper, all models perform near chance across all ICL strategies and shot counts, with no consistent model advantage on fairness metrics.

\begin{table*}[t]
\centering
\caption{Accuracy and fairness metrics under four ICL strategies evaluated on the COMPAS recidivism dataset with race as the protected attribute. Higher accuracy is better; lower values indicate improved fairness for DPD, EoD, and EoOpp.}
\resizebox{\textwidth}{!}{%
\begin{tabular}{ll cccc|cccc|cccc}
\toprule
& & \multicolumn{4}{c}{VaultGemma-1B} & \multicolumn{4}{c}{Gemma-3-1B-PT} & \multicolumn{4}{c}{Gemma-2-2B} \\
\cmidrule(lr){3-6} \cmidrule(lr){7-10} \cmidrule(lr){11-14}
Selection & $k$ & Acc & DPD & EoD & EoOpp
& Acc & DPD & EoD & EoOpp
& Acc & DPD & EoD & EoOpp \\
\midrule
\multirow{3}{*}{Random}
& 0 & 49.67 & 0.05 & 0.06 & 0.03 & 44.00 & 0.50 & 0.55 & 0.55 & 50.00 & 0.00 & 0.00 & 0.00 \\
& 4 & 50.17 & 0.02 & 0.03 & 0.01 & 52.50 & 0.02 & 0.03 & 0.03 & 53.17 & 0.06 & 0.08 & 0.08 \\
& 8 & 50.50 & 0.00 & 0.01 & 0.00 & 51.83 & 0.00 & 0.01 & 0.00 & 52.33 & 0.05 & 0.07 & 0.07 \\
\midrule
\multirow{3}{*}{Balanced}
& 0 & 49.67 & 0.05 & 0.06 & 0.03 & 44.00 & 0.50 & 0.55 & 0.55 & 50.00 & 0.00 & 0.00 & 0.00 \\
& 4 & 54.00 & 0.05 & 0.08 & 0.08 & 54.33 & 0.09 & 0.15 & 0.15 & 58.67 & 0.23 & 0.29 & 0.29 \\
& 8 & 50.00 & 0.01 & 0.01 & 0.01 & 56.67 & 0.14 & 0.19 & 0.19 & 52.17 & 0.08 & 0.13 & 0.13 \\
\midrule
\multirow{3}{*}{Instruction}
& 0 & 50.00 & 0.00 & 0.00 & 0.00 & 45.00 & 0.07 & 0.07 & 0.07 & 44.33 & 0.37 & 0.41 & 0.32 \\
& 4 & 51.00 & 0.03 & 0.03 & 0.03 & 51.83 & 0.02 & 0.03 & 0.03 & 53.67 & 0.07 & 0.07 & 0.07 \\
& 8 & 50.00 & 0.01 & 0.02 & 0.01 & 52.50 & 0.00 & 0.03 & 0.02 & 51.83 & 0.06 & 0.10 & 0.10 \\
\midrule
\multirow{3}{*}{Removal}
& 0 & 49.67 & 0.01 & 0.01 & 0.00 & 51.17 & 0.04 & 0.05 & 0.02 & 50.00 & 0.00 & 0.00 & 0.00 \\
& 4 & 51.00 & 0.03 & 0.03 & 0.03 & 52.00 & 0.03 & 0.07 & 0.07 & 52.50 & 0.01 & 0.02 & 0.00 \\
& 8 & 50.17 & 0.01 & 0.01 & 0.01 & 51.17 & 0.00 & 0.01 & 0.01 & 53.50 & 0.03 & 0.03 & 0.03 \\
\bottomrule
\end{tabular}%
}
\label{tab:fairness_results_compas}

\vspace{6pt}

\centering
\caption{Accuracy and fairness metrics under four ICL strategies evaluated on the German Credit dataset with sex as the protected attribute. Higher accuracy is better; lower values indicate improved fairness for DPD, EoD, and EoOpp.}
\resizebox{\textwidth}{!}{%
\begin{tabular}{ll cccc|cccc|cccc}
\toprule
& & \multicolumn{4}{c}{VaultGemma-1B} & \multicolumn{4}{c}{Gemma-3-1B-PT} & \multicolumn{4}{c}{Gemma-2-2B} \\
\cmidrule(lr){3-6} \cmidrule(lr){7-10} \cmidrule(lr){11-14}
Selection & $k$ & Acc & DPD & EoD & EoOpp & Acc & DPD & EoD & EoOpp & Acc & DPD & EoD & EoOpp \\
\midrule
\multirow{3}{*}{Random} & 0 & 50.00 & 0.00 & 0.00 & 0.00 & 50.33 & 0.01 & 0.01 & 0.01 & 53.33 & 0.12 & 0.15 & 0.15 \\
& 4 & 49.33 & 0.00 & 0.00 & 0.00 & 55.00 & 0.11 & 0.12 & 0.11 & 50.33 & 0.10 & 0.11 & 0.09 \\
& 8 & 0.00 & 0.00 & 0.00 & 0.00 & 51.67 & 0.06 & 0.09 & 0.09 & 53.00 & 0.05 & 0.07 & 0.03 \\
\midrule
\multirow{3}{*}{Balanced} & 0 & 50.00 & 0.00 & 0.00 & 0.00 & 50.33 & 0.01 & 0.01 & 0.01 & 53.33 & 0.12 & 0.15 & 0.15 \\
& 4 & 49.00 & 0.02 & 0.04 & 0.00 & 51.67 & 0.06 & 0.09 & 0.09 & 51.00 & 0.02 & 0.04 & 0.00 \\
& 8 & 0.00 & 0.00 & 0.00 & 0.00 & 54.00 & 0.03 & 0.07 & 0.07 & 50.33 & 0.01 & 0.01 & 0.01 \\
\midrule
\multirow{3}{*}{Instruction} & 0 & 52.00 & 0.12 & 0.17 & 0.17 & 50.00 & 0.00 & 0.00 & 0.00 & 50.00 & 0.00 & 0.00 & 0.00 \\
& 4 & 49.67 & 0.02 & 0.03 & 0.01 & 54.33 & 0.10 & 0.11 & 0.11 & 51.00 & 0.11 & 0.13 & 0.13 \\
& 8 & 0.00 & 0.00 & 0.00 & 0.00 & 51.67 & 0.10 & 0.12 & 0.12 & 53.33 & 0.03 & 0.04 & 0.01 \\
\midrule
\multirow{3}{*}{Removal} & 0 & 50.00 & 0.00 & 0.00 & 0.00 & 50.00 & 0.00 & 0.00 & 0.00 & 53.67 & 0.01 & 0.12 & 0.11 \\
& 4 & 50.00 & 0.00 & 0.01 & 0.01 & 52.33 & 0.05 & 0.12 & 0.03 & 51.67 & 0.02 & 0.03 & 0.01 \\
& 8 & 0.00 & 0.00 & 0.00 & 0.00 & 50.33 & 0.02 & 0.03 & 0.03 & 53.67 & 0.01 & 0.03 & 0.03 \\
\bottomrule
\end{tabular}%
}
\label{tab:fairness_results_german}
\end{table*}

\end{document}